\documentclass[12pt]{amsart}
 
\usepackage[margin=1in]{geometry} 
\usepackage{amsmath,amsthm,amssymb}
\usepackage{graphicx}
\usepackage{algorithm}
\usepackage[noend]{algpseudocode}

\newcounter{question}[section]

\title{Gravilon: Applications of a New Gradient Descent Method to Machine Learning}

\author{Chad Kelterborn}
\author{Marcin Mazur}
\author{Bogdan V.~Petrenko}

\address{
Department of Mathematics \\
University of North Carolina - Chapel Hill \\
120 E Cameron Avenue \\
CB \#3250 \\
329 Phillips Hall \\
Chapel Hill, NC 27599, USA
}
\email{ckelter@live.unc.edu}

\address{
Department of Mathematics \\
Binghamton University \\
P.O. Box 6000 \\
Binghamton, NY 13892-6000, USA } 
\email{mazur@math.binghamton.edu}

\address{
Department of Mathematics and Computer Science\\ 
Eastern Illinois University \\ 
600 Lincoln Avenue \\ 
Charleston, IL 61920-3099, USA
}
\email{bvpetrenko@eiu.edu}

\begin{document}

\maketitle

\section{Abstract}
Gradient descent algorithms have been used in countless applications since the inception of Newton's method. The explosion in the number of applications of neural networks has re-energized efforts in recent years to improve the standard gradient descent method in both efficiency and accuracy. These methods modify the effect of the gradient in updating the values of the parameters. These modifications often incorporate hyperparameters: additional variables whose values must be specified at the outset of the program. We provide, below, a novel gradient descent algorithm, called Gravilon, that uses the geometry of the hypersurface to modify the length of the step in the direction of the gradient. Using neural networks, we provide promising experimental results comparing the accuracy and efficiency of the Gravilon method against commonly used gradient descent algorithms on MNIST digit classification. \\

\textbf{Keywords}: gradient descent, neural networks, adaptive learning rates, numerical methods, machine learning, optimization

\section{Introduction}
This is a preliminary report of ongoing research. We believe that this report is warranted because of the promising experimental results. We welcome comments or suggestions for improvements. \\

Gradient descent methods readily find uses in any application that requires one to update parameters. In computer programming, examples include optimization methods, neural networks, and deep learning. Gradient descent provides an algorithmic process for finding local extrema of a function from the gradient. From a historical perspective, the most well known method is the one often attributed to Newton.  \\

Consider the graph of the function $f(x)$, where $x$ is a real variable, and suppose that we want to find a local minimum via gradient descent using Newton's method. The process begins by first choosing an initial point, say $x_0$, and continues by describing an iterative method on how to obtain each subsequent point from the previous. This is done in the following manner \\

$$x_{t+1} = x_t - f'(x_t)$$
One obtains the $(t+1)$st term in the descent sequence from the $(t)$th term by subtracting from the $(t)$th term the value of the derivative of the function evaluated at the $(t)$th term. \\

This process easily extends to a function of $n$ real variables. Let $\textbf{x}$ be shorthand notation for the $n$ variables $x_1, \ldots, x_n$. We use the convention that the $(t)$th value of the variable $x_i$ has the notation $x_{i,t}$ throughout. Then using Newton's method one obtains the $(t+1)$st point in the descent from the $(t)$th point as

$$\textbf{x}_{t+1} = \textbf{x}_t - \nabla f(\textbf{x}_t)$$

After some thought, one may wonder if the distance of travel for each step prescribed by Newton's method is optimal. The fact that the gradient of a function provides only an optimal direction in the neighborhood of a point lends credence to this idea. The question then becomes, how can one change the length of the step in the direction of the gradient to improve upon both the efficiency and accuracy of Newton's method? Attempts to answer this question have been developed in the world of neural networks where exotic gradient descent algorithms, such as Adam, Adagrad, and RMSProp, modify the effect of the gradient when updating variables. These modifications rely on some combination of additional information such as introducing new parameters (called hyperparameters), introducing some type of decay, or incorporating some function of the values of previous gradients -  which could mean all of them, a moving average, or just the previous gradient. In the realm of neural networks, each of these gradient descent algorithms has accomplished, one way or another, the goal of improving the efficiency and accuracy of Newton's method. \\

We propose a new gradient descent method. The Gravilon method uses the geometry of the hypersurface to modify the length of the step in the direction of the gradient. We conduct experiments comparing the efficiency and accuracy of the Gravilon method against gradient descent algorithms commonly employed in neural networks. In our experiments, the gradient descent algorithms that we consider include the standard SGD, Adagrad, Adam, Adamax, Nadam, and RMSprop. Unlike these algorithms, the Gravilon method does not make use of any hyperparameters. So, one does not need to manually specify the global learning rate nor is hyperparameter tuning needed. Furthermore, this method has the additional benefit that it can be deployed in conjunction with the exotic gradient descent algorithms that are commonly used in neural networks. This hints at the possibility of the existence of a hybrid method. \\

\textbf{Acknowledgements}: Rarely does a project precipitate in a complete vacuum, devoid of any external support. This is not one of those rarities. Without the guiding help and support of numerous individuals, this project would not have been possible. Notably, we would like to express our deepest gratitude to James Choi, Trevor Kelterborn, Richard Koss, Abidalrahman Moh'd, and Daniel Paydarfar. \\

\section{Gradient Descent Method}
In this section, we describe the geometric idea behind the Gravilon method for gradient descent. We extend the method that was developed in the master's thesis written by Richard Koss under the supervision of Bogdan Petrenko \cite{koss}. \\

\subsection{Gravilon Method} 
Let's consider a function $f: \mathbb{R}^n \longrightarrow \mathbb{R}$ with $z = f(x_1, \ldots, x_n)$ with the restriction that the function can be made nonnegative definite by a constant shift such that minima of $f(x_1, \ldots, x_n)$ have height $0$. In its current state, the method that we propose below takes advantage of the fact that we know what the value of the minimum of the function $f(x_1, \ldots, x_n)$ is and that we are only looking for where the minimum occurs. \\

Suppose that starting at the point $P:(a_1, \ldots, a_n, f(a_1, \ldots,a_n))$ on the hypersurface $z = f(x_1, \ldots, x_n)$ we would like to follow the path of steepest descent - namely, a path in the negative of the direction of the gradient - to a local extremum. We describe an iterative method that produces a sequence of points converging to this extremum. Starting from the point $(a_1,\ldots,a_n)$, we need to describe how to obtain the next point in the sequence. The first step is to compute the gradient of the function $G(x_1,\ldots,x_n,z) = f(x_1,\ldots,x_n) - z$ :
\begin{equation}
    \label{step1}
    \nabla G(x_1,\ldots,x_n,z) = \langle f_{x_1}(x_1,\ldots,x_n), \ldots, f_{x_n}(x_1,\ldots,x_n), -1 \rangle
\end{equation}
where $f_{x_i}(x_1,\ldots,x_n)$ is the $i$th partial derivative of $f$.\\

Recall that the gradient vector of a function produces a normal vector to the hypersurface based at a specified point. From the normal vector we can construct the hyperplane tangent to the hypersurface $z=f(x_1,...,x_n)$ based at the point $P$, and let's call this hyperplane $T_P$. The equation of this hyperplane is given by
\begin{equation}
    \label{tangentplane}
    T_P: f_{x_1}(a_1,\ldots,a_n)(x_1 - a_1) + \ldots + f_{x_n}(a_1,\ldots,a_n)(x_n - a_n) - (z-f(a_1, \ldots,a_n)) = 0
\end{equation}

The second step is to project the gradient vector $\nabla G(a_1,\ldots,a_n,z)$ into the $z = 0$ hyperplane and to construct the line in the direction of the resulting vector passing through the point $(a_1,\ldots,a_n,0)$. In parametric form, the equation of this line is given by
\begin{equation}
    \label{line}
    l(s) = s \langle f_{x_1}(a_1,\ldots,a_n), \ldots, f_{x_n}(a_1,\ldots,a_n), 0  \rangle + \langle a_1, \ldots, a_n, 0 \rangle
\end{equation}

The third step is to find the point where the line $l(s)$ intersects the tangent plane $T_P$. This point is found by substituting $s f_{x_i}(a_1,\ldots,a_n) + a_i$ for $x_i$ in $T_P$ and solving for $s$. In this computation, when we write the partial derivative $f_{x_i}$, we really mean the partial derivative evaluated at the point $(a_1,\ldots,a_n)$, $f_{x_i}(a_1,\ldots,a_n)$. 
\begin{center}
    \label{intersection}
    $T_P: f_{x_1}(x_1 - a_1) + \ldots + f_{x_n}(x_n - a_n) - (z-f(a_1, \ldots,a_n)) = 0$ \\[15pt]
    $f_{x_1}(s f_{x_1}) + \ldots + f_{x_n}(s f_{x_n}) - (0 - f(a_1, \ldots,a_n))  = 0$ \\[15pt]
    $s = \dfrac{-f(a_1, \ldots,a_n)}{f_{x_1}^2 + \ldots + f_{x_n}^2}$ \\[15pt]
    $s = \dfrac{-f(a_1, \ldots,a_n)}{|\nabla f(a_1, \ldots, a_n)|^2}$
\end{center}

Substituting for $s$ in the equation of the line $l(s)$, we obtain the point

\begin{center}
    \label{point}
    $l\left(\dfrac{-f(a_1, \ldots,a_n)}{|\nabla f(a_1, \ldots, a_n)|^2}\right) = \dfrac{-f(a_1, \ldots,a_n)}{|\nabla f(a_1, \ldots, a_n)|^2} \langle f_{x_1}, \ldots, f_{x_n},0 \rangle + \langle a_1, \ldots, a_n, 0 \rangle$ \\[15pt]
\end{center}

This point is the next point in our sequence. Projecting this point onto the hypersurface $z = f(x_1, \ldots, x_n)$, we then repeat this process to produce a sequence converging to a local extremum. To obtain the $(t+1)$st term of the sequence from the $(t)$th term we use the formula 

\begin{center}
    $(a_{1,t+1}, \ldots, a_{n,t+1}) = (a_{1,t}, \ldots, a_{n,t}) - \dfrac{f(a_{1,t}, \ldots,a_{n,t})}{|\nabla f(a_{1,t}, \ldots, a_{n,t})|^2} (f_{x_1}, \ldots, f_{x_n})$
\end{center}
where again the partial derivatives $f_{x_i}$ are evaluated at the point $(a_{1,t}, \ldots, a_{n,t})$, $f_{x_i} = f_{x_i}(a_{1,t}, \ldots, a_{n,t})$, or in condensed notation with $\textbf{a}_{t} = (a_{1,t}, \ldots, a_{n,t})$

$$\textbf{a}_{t+1} = \textbf{a}_{t} - \dfrac{f(\textbf{a}_{t})}{|\nabla f(\textbf{a}_{t})|^2} \nabla f(\textbf{a}_{t})$$

The benefits of this method are at least two-fold. One is not restricted in one's choice of the initial descent point. Indeed, the wealth of success that this method has enjoyed in experiments involving neural networks as well as in other applications strongly suggests that this is a global method because it converges to a local minimum regardless of the initial descent point. Additionally, the method converges at least as fast as the standard gradient descent method. In fact, we have a proof of convergence near a local minimum. The rate of this convergence is exponential. \\

At this time, however, we do not have a thorough theoretical analysis of the convergence of this method. As we mentioned earlier, the method works for nonnegative definite functions whose minima have height $0$. In order for the method to converge to a minimum, we need the term $$\dfrac{f(\textbf{a}_{t})}{|\nabla f(\textbf{a}_{t})|^2} \nabla f(\textbf{a}_{t})$$ 
to converge to $0$. Roughly speaking, the function value $f(\textbf{a}_{t})$ has to at least be less than $\epsilon$ if we say that the magnitude of the gradient $|\nabla f(\textbf{a}_{t})|^2 = \epsilon^2$. Some of the challenges that we face are similar to those of Newton's method. The paper \cite{hubbard} explores some of these challenges as they pertain to Newton's method.\\

\section{Neural Networks}
In this section, we provide a brief account of neural networks. If you feel that you are already familiar with neural networks, please feel free to skip this section and move onto the results in Section \ref{experiments}. On the other hand, for further details we have found the following resources helpful \cite{nielsen}, \cite{ruder2016overview}, \cite{veen}, and \cite{goodfellow}. \\

\subsection{Example}
Let's imagine for a moment that we are tasked with the problem of classifying handwritten digits from $0$ to $9$. If the data set of images is reasonably large, say $60,000$ images, such a task quickly becomes onerous trying to do it by hand. In an effort to make our lives easier, we decide to create a program that will complete the classification autonomously. This program should act as a function, $F$, that reads in an image of a handwritten digit and outputs a value from $0$ to $9$ - the function's determination of the value of the handwritten digit. \\

The input image, which let's suppose is a $28 \times 28$ pixel grayscale image, can be quantified in terms of the values of the pixels of the image. For a grayscale image, the value of a pixel quantifies the brightness of the pixel: typically, $0$ corresponds to black and $255$ to white. Then to each image we can associate a $784 \times 1$ column vector where each entry corresponds to the value of a pixel of the grayscale image, normalized by dividing out by $255$ to control the size of the gradients in later computations. After applying the function $F$ to a particular input vector, we obtain a $10 \times 1$ vector probability distribution, where the $i$th component, $i = 0,1, \ldots, 9$, assigns the probability that the image is the $i$th digit. The largest probability in this vector will determine the digit to which the image is assigned. \\

We now need a way to compare the computed classification and the image's actual classification. For each image in our data set, let's assume that there is a label that correctly identifies the handwritten digit. Such an assumption is not far-fetched as at this stage we train our function via supervised learning. We transform this label into a $10 \times 1$ column vector, denoted $\textbf{y}$, where the only nonzero entry is a $1$ in the $i$th entry. Using the cross entropy metric from probability theory, we can quantify the discrepancy between the function's classification of the image and the actual classification. Recall that if two discrete probability distributions $p$ and $q$ have the same support $\mathcal{F}$, then one computes the cross entropy of $p$ and $q$ as  \\

\begin{equation}
    H(p,q) = -\sum_{x \in \mathcal{F}} p(x) \log(q(x))
\end{equation}

The metric is more commonly referred to as a loss or cost function. Often it is a nonnegative definite function, and the smaller the value of the loss function, we say the better the function $F$ performed at classifying the input image. The goal is to find a function $F$ that minimizes the cross entropy loss function. 

\begin{figure}[ht]
\includegraphics[scale=0.75]{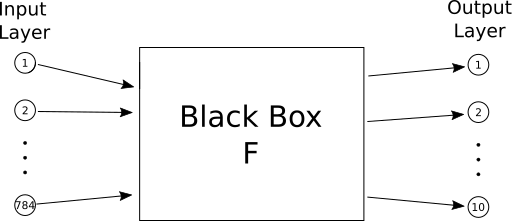}
\centering
\caption{Black box diagram of Neural Network}
\label{fig:blackbox}
\end{figure}

\begin{figure}[ht]
\includegraphics[scale=0.75]{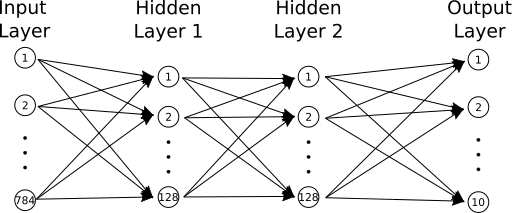}
\centering
\caption{Neural network with two hidden layers each consisting of 128 nodes.}
\label{fig:NN_2hl}
\end{figure}

At this stage, our program can be summarized according to the Figure (\ref{fig:blackbox}), we feed a column vector of the normalized values of a grayscale image into a mysterious black box that outputs a probability distribution from which the image is classified. There are many choices that we can make regarding the structure of the black box, and in the next section we will discuss some of these choices. In an effort to remove some of the mystery, the black box represents compositions of functions, namely affine transformations and activation functions applied componentwise, see equations (\ref{sigmoid}) and (\ref{softmax}) below for examples of activation functions. For now, let's consider a black box with two hidden layers, aptly named hidden layer $1$ and hidden layer $2$, where hidden layer $1$ has $128$ nodes and hidden layer $2$ has $128$ nodes. The network has the form depicted in the Figure (\ref{fig:NN_2hl}). Nodes represent entries of a column vector. Arrows represent the matrix multiplication, $A_i \textbf{v}$, component of the affine transformations. Said otherwise, the arrows show which inputs affect which outputs. \\

Deciding upon this structure of a neural network immediately determines that we will successively compose three pairs of affine transformations and activation functions. The affine transformations have the form $A_i \textbf{v} + \textbf{b}_i$ for $i = 1,2,3$. One can think of a hidden layer as an intermediate output obtained from the composition of an affine transformation and activation function applied to an input vector $\textbf{v}$. That is if the input layer is understood to be the $0$th layer, then the output of the $i$th layer is $act(A_i \textbf{v} + \textbf{b}_i)$ where $act$ denotes the activation function. The activation function is often a real-valued function of a single real variable. So, when we apply the activation function to a vector $\textbf{v}$, we embrace the notation $act(\textbf{v})$ to mean that the activation function is applied to each component of the vector $\textbf{v}$. \\

In general, the matrix $A_i$ has dimensions $(\text{nodes in layer } i) \times (\text{nodes in layer } i-1)$ and the column vector $\textbf{b}_i$ has dimensions $(\text{nodes in layer } i) \times 1$. The entries in the matrices $A_1$, $A_2$, and $A_3$, called weights, and the entries in the column vectors $\textbf{b}_1$, $\textbf{b}_2$, and $\textbf{b}_3$, called biases, comprise the variables that will be updated to minimize the loss function. We will use the notation that $a^i_{jk}$ represents the $(jk)$th entry in the matrix $A_i$ and $b^i_j$ represents the $j$th entry of the column vector $\textbf{b}_i$. When we say that we try to optimize the neural network, we mean that we look for values of the weights and biases that approximate a minimum of the loss function. \\

When optimizing our neural network (function), care must be taken to ensure that the optimized function has not been designed to solely recognize characteristics of the input images. A successful function should have the ability to accurately classify images of handwritten digits not included in the data set. One way to reduce this risk is to create two subsets of the $60,000$ images: a training data set consisting of $50,000$ randomly selected images, and a testing data set consisting of the remaining $10,000$ images. We use the training data set to update the variables and optimize the function, then we determine the function's accuracy by using images from the testing data set. \\

Using all $50,000$ images in the training data set to optimize the function can be computationally time consuming. Instead we use subsets, called batches, to train the neural network. There is something to be said, however, about the accuracy of the neural network in the limit of the number of elements in the training data set. One surmises that the larger the training data set, the better the performance of the neural network; though, at this time, this statement lacks rigorous proof. \\

Let $\mathcal{B} = \{x^{(1)}, \ldots, x^{(N)}\}$ denote a training batch subset whose size equals the number of elements in the set, $|\mathcal{B}| = N$. Consider an image $x$ from our training batch $\mathcal{B}$, we take the associated $784 \times 1$ column vector, $\textbf{v}_{x}$, and feed it through the network and compare the computed probability distribution with the actual probability distribution, $\textbf{y}_{x}$, arising from the image's label. \\

The neural network first applies an affine transformation to the input vector $\textbf{v}_x$. We obtain the vector $A_1 \textbf{v}_x + \textbf{b}_1$. To this vector we apply the sigmoid activation function (\ref{sigmoid}) componentwise. The vector $\textbf{z}_{x,1} = \sigma(A_1 \textbf{v}_x + \textbf{b}_1)$ is the output of hidden layer $1$. Explicitly, the vector

\begin{equation*}
    A_1 \textbf{v}_x + \textbf{b}_1 = 
    \begin{pmatrix}
        \sum_k a_{1k}^1 v_{x,k} + b_1^1 \\
        \vdots \\
        \sum_k a_{128k}^1 v_{x,k} + b_{128}^1
    \end{pmatrix}
\end{equation*}
transforms under the activation function $\sigma$ as

\begin{equation*}
    \textbf{z}_{x,1} = \sigma(A_1 \textbf{v}_x + \textbf{b}_1) = 
    \begin{pmatrix}
        \sigma(\sum_k a_{1k}^1 v_{x,k} + b_1^1) \\
        \vdots \\
        \sigma(\sum_k a_{128k}^1 v_{x,k} + b_{128}^1)
    \end{pmatrix}
\end{equation*}
A similar computation yields the outputs of hidden layer $2$ and the output layer. The vector $\textbf{z}_{x,2} = \sigma(A_2 \textbf{z}_{x,1}+ \textbf{b}_2)$ is the output of hidden layer $2$. The vector $\textbf{z}_{x,3} = \varphi(A_3 \textbf{z}_{x,2} + \textbf{b}_3)$ is the output of the output layer where $\varphi$ denotes the softmax function (\ref{softmax}). \\

The sigmoid function, denoted $\sigma(z)$, is a real function of a single real variable given by the equation

\begin{equation}\label{sigmoid}
    \sigma(z) = \dfrac{1}{1+e^{-z}}    
\end{equation}
We apply the sigmoid function to a vector componentwise. The softmax function is a vector valued function, and in our case it is given by $\varphi(\textbf{z}) : \mathbb{R}^{10} \longrightarrow \mathbb{R}^{10}$

\begin{equation}\label{softmax}
    \varphi(\textbf{z})_i = \dfrac{e^{z_i}}{\sum_{j=1}^{10}e^{z_j}}    
\end{equation}
for $i = 1, \ldots, 10$ and $\textbf{z} = (z_1, \ldots, z_{10})$. Note that the softmax function returns a vector such that all of its entries lie in the interval $(0,1)$ and the sum of all of its entries is $1$. The resulting vector can thus be interpreted as a probability distribution over the predicted output categories. With this probability distribution, we compute the loss function according to the cross entropy metric:

\begin{equation}
    Loss(\textbf{z}_{x,3}, \textbf{y}_x) = \sum_{i=1}^{10} -\textbf{y}_x^i \log(\textbf{z}_{x,3}^i)    
\end{equation}

We repeat this process for each element in the training batch, $\mathcal{B}$, and average the loss values over each element in the training batch. It is this function/quantity that we will minimize

\begin{equation}
    Loss(a^i_{jk},b^i_{j}) = \frac{1}{N}\sum_{x \in \mathcal{B}} Loss(\textbf{z}_{x,3}, \textbf{y}_x)
\end{equation}
where $N$ denotes the batch size. We express the loss function in terms of the parameters to be updated, and we use $a^i_{jk}$ and $b^i_{j}$ as shorthand notation for all of the weights and biases in the neural network. \\

The last step is to compute the gradient of the loss function and update the variables through a method called backpropagation, a method first discussed in \cite{rumelhartbackprop}, which makes use of the chain rule to update parameters. The observation to make here is that the partial derivatives of the variables in earlier layers of the network are expressed in terms of the partial derivatives of the variables in later layers of the network. By first computing the partial derivatives of the loss function with respect to the variables $a^3_{jk}$ and $\textbf{b}^3_j$, one then can compute the partial derivatives of the loss function with respect to $a^2_{jk}$ and $\textbf{b}^2_j$, and then with respect to $a^1_{jk}$ and $\textbf{b}^1_j$. Diagrammatically, we propagate backwards through the neural network as we compute the partial derivatives. Once we have computed the partial derivatives, we update the parameters according to the chosen gradient descent method. \\

Note that only one of the coordinates of the vector $\textbf{y}_x$ is nonzero. So for a particular input $\textbf{x}$, only one term in the summation contributes to the loss function $Loss(\textbf{z}_{x,3}, \textbf{y}_x)$. Since the loss function is defined as the negative of the logarithm of a probability distribution, the loss function is nonnegative definite and the minimum of the function has height $0$. Following the Gravilon method to update the parameters, we see that we obtain the $(t+1)$st value of the weight $a^i_{jk}$ from the $(t)$th value as

\begin{equation}
    a^i_{jk,t+1} = a^i_{jk,t} - \dfrac{Loss(a^i_{jk,t}, b^i_{j,t})}{|\nabla Loss(a^i_{jk,t}, b^i_{j,t})|^2} Loss_{a^i_{jk}}(a^i_{jk,t}, b^i_{j,t})    
\end{equation}
and similarly, we obtain the $(t+1)$st value of the bias $\textbf{b}^i_j$ from the $(t)$th value as

\begin{equation}
    b^i_{j,t+1} = b^i_{j,t} - \dfrac{Loss(a^i_{jk,t}, b^i_{j,t})}{|\nabla Loss(a^i_{jk,t}, b^i_{j,t})|^2}Loss_{b^i_j}(a^i_{jk,t}, b^i_{j,t})    
\end{equation}
The subscript $a^i_{jk}$ on the loss function denotes the partial derivative of the loss function with respect to the variable $a^i_{jk}$. Similarly for $b^i_j$. We repeat this process until the loss function has attained a minimum. Algorithm \ref{alg:Gravilon} shows how to follow the gradient descent according to the Gravilon method.

\begin{algorithm}
    \caption{Gravilon gradient descent update}\label{alg:Gravilon}
    \begin{algorithmic}
        \State \textbf{Require:} initial parameter $\theta$
        \While{stopping criteria not met}
        \State Sample a minibatch of $N$ examples from the training set $\{x^{(1)}, \ldots, x^{(N)}\}$ with 
        \State corresponding targets $y^{(i)}$.
        \State Compute the gradient estimate: $g \leftarrow \frac{1}{N}\nabla_{\theta} \sum_i L(f(x^{(i)}; \theta), y^{(i)})$
        \State Compute magnitude squared of gradient: $|g|^2 \leftarrow g \cdot g$
        \State Compute step size update: $\beta = \frac{\frac{1}{N}\sum_i L(f(x^{(i)}; \theta), y^{(i)})}{|g|^2}$
        \State Apply update: $\theta \leftarrow \theta - \beta g$
        \EndWhile \label{Gravilonendwhile}
        \State \textbf{end while}
    \end{algorithmic}
\end{algorithm}

\subsection{General}
In this section, we propose a generalization of some of the ideas that were discussed in the previous section. By no means do we expect this to be a complete account of neural networks. For additional information and other viewpoints, please consider reviewing some of the many resources available on neural networks such as the book \cite{nielsen}. \\

A neural network is a function $f: \mathbb{R}^n \longrightarrow \mathbb{R}^k$ that can be used to classify data into predetermined categories. If we concentrate our efforts momentarily on image recognition, then we readily see some examples from the basics of classifying handwritten digits and recognizing road signs to complicated processes of object recognition and path-planning in self-driving vehicles. There are many vector valued functions from which we can choose; however, we are most interested in choosing the best function, or at least a good approximation of the best function, that accurately classifies the data. Over the space of functions, our goal is to optimize the function according to a metric, often called a loss function, that we are free to choose. Finding the minimizing function over the space of all vector valued functions $f:\mathbb{R}^n \longrightarrow \mathbb{R}^k$ quickly proves to be an intractable problem. \\

Instead, we restrict our focus to a certain class of functions that can be described as compositions of affine transformations and activation functions applied componentwise. In the previous section, we saw two examples of activation functions: the sigmoid function $(\ref{sigmoid})$ and the softmax function $(\ref{softmax})$. It has been shown in \cite{cybenko} that this class of functions can approximate any continuous function. While this result allows us to narrow our focus to a particular class of functions, a choice of a function still has to be made for each data set. The prescribed method for making an initial choice of such a function and then optimizing over the function space is trial and error. \\

Some characteristics of the neural network are determined by the data. Namely, 
the presentation of the data to be classified determines the number of nodes (components) in the input layer, and the number of preassigned categories in which the data can be classified determines the number of nodes in the output layer. What remains to be decided is the structure of the hidden layers of the neural network. Here we have great flexibility and even a chance for further exploration as a clear understanding of the function and construction of hidden layers remains elusive at this time. \\

In the previous section, we constructed an example of a neural network consisting of two hidden layers which we viewed as intermediate steps in a sequence of compositions of affine transformations and activation functions. This is an example of a multilayer perceptron neural network, and it is, probably, the most basic structure of the hidden layers we can construct. We readily see how this example may be extended to more layers - corresponding to the compositions of more affine transformations and activation functions - and we see that varying the number of nodes in each layer corresponds to varying the dimensions of the affine transformation. When constructing a neural network, even if we restrict ourselves to this multilayer perceptron architecture, it is not etched in stone how to decide the number of layers and the number of nodes per layer. \\

A wide variety of neural networks have been developed to tackle different problems, convolutional neural networks have been employed in questions involving computer vision and recurrent neural networks have been employed to answer questions revolving around time-based data predictions. For a detailed discussion on these types of neural networks (and others) please consider looking at \cite{veen} and the references therein. \\

Having decided upon a design for the neural network and chosen which metric (loss function) will be used, we must then update the parameters of the neural network, a process traditionally done via gradient descent. In gradient descent, one obtains the $(t+1)$st value of the parameter $x_i$ of the loss function from the $(t)$th value as

\begin{equation}
    x_{i,t+1} = x_{i,t} - loss_{x_i}(\textbf{x}_{t})
\end{equation}
where $loss_{x_i}$ denotes the $i$th partial derivative of the loss function. Furthermore, we denote the collection of arguments of the $loss$ function as $\textbf{x}$ as the number of arguments can be any number. \\

Work has been done in modifying the standard gradient descent algorithm to improve the efficiency and accuracy of the algorithm. These modifications alter the amount that the partial derivative $loss_{x_i}$ affects the update of the parameter $x_i$. The simplest modification is the introduction of a constant coefficient on the partial derivative

\begin{equation}
    x_{i,t+1} = x_{i,t} - c \cdot loss_{x_i}(\textbf{x}_{t})
\end{equation}
where $c$ is some positive real constant, usually $c \sim 0.001$. More exotic modifications have been created that take into account the values of some or all past gradients as well as introducing hyperparameters, which are additional variables whose values must be specified before running the program. With no prescribed method for divining the optimal values of these hyperparameters available, which have to be fine tuned to the question at hand, we quickly see that we have increased the complexity of finding the optimal neural network. For a discussion on various gradient descent optimization algorithms as well as a discussion on other strategies for optimizing gradient descent, we found the paper \cite{ruder2016overview} illuminating. \\

We say that a neural network is ``trained'' when the values of the parameters that minimize the loss function have been obtained. There is some flexibility in deciding when a network is trained. Early stopping, a choice largely made by the programmer, can be decided by a variety of thresholds, for example the number of iterations the program runs or a plateauing of the classification accuracy. It should be noted that if the program runs for too long then the program is susceptible to over-fitting the data. \\

The one dimensional case of fitting $n$ points plotted in the $xy$-coordinate plane can help us visualize this idea. Generally speaking, for $n$ arbitrary points in the plane there is a unique $(n-1)$-degree polynomial, $f(x)$, passing through all of the points. The question then becomes do we choose this polynomial $f(x)$ or the line of best fit $y = mx + b$ as the function that best fits the data? To answer this question, we would need more information about the data that we are trying to fit. It might be reasonable to assume that the line is in fact the best approximation of the data as it would be strange for the data to exhibit linear behavior without some underlying truth to it, noting that the points' deviation from this line is noise or error from the experiment. Continuing this rationale, we see that the polynomial $f(x)$, although it exactly fits all of the points, takes into account local noise when fitting the data. Generalizing this example, a neural network that runs too long may begin to recognize local noise when optimizing the parameters. In recognizing more subtle features of the training data, the neural network loses the ability to recognize overarching trends diminishing its ability to accurately classify general data. \\

\section{Experiments}
\label{experiments}
In the following experiments, we consider a variety of gradient descent algorithms for MNIST digit classification. We hope that the results serve as a springboard, encouraging further experimentation of the gradient descent algorithm, Gravilon, applied to other classification problems. Currently, several theoretical aspects of the Gravilon method remain a mystery. Not the least of which is the convergence of this method. 

\subsection{MNIST Digit Classification}
In this first set of experiments, we compared the classification accuracy of the Gravilon method against the standard SGD \cite{robbinssgd}, Adagrad \cite{duchiadagrad}, Adam \cite{kingma2014adam}, Adamax \cite{kingma2014adam}, Nadam \cite{dozatnadam}, and RMSprop \cite{hintonrmsprop} gradient descent algorithms for the MNIST handwritten digit classification \cite{lecun}. We completed $15$ trials for each method where each trial runs for $2500$ steps. In Table \ref{tab:Trial1}, we present the average classification accuracy and the best accuracy over the $15$ trials for each method. \\ 

The architecture of the neural network that we used has two hidden layers, the first containing $128$ nodes and the second containing $128$ nodes. We used the sigmoid activation function in hidden layers $1$ and $2$. We trained the neural network using stochastic gradient descent (SGD), normalizing the pixel values to the interval $[0,1]$, and clipping the gradient with a threshold value of $1$. We set the mini batch size to $256$. For these experiments, we used the following values of the hyperparameters for each gradient descent algorithm: \\

\textbf{SGD} learning rate $\alpha = 0.001$ \\

\textbf{Adagrad} learning rate $\alpha = 0.001, 0.01, 0.1$, initial accumulator value $= 0.1$, $\epsilon = 1\mathrm{e}{-7}$ \\

\textbf{Adam} learning rate $\alpha = 0.001$, $\beta_1 = 0.9$, $\beta_2 = 0.999$, $\epsilon = 1\mathrm{e}{-7}$ \\

\textbf{Adamax} learning rate $\alpha = 0.001$, $\beta_1 = 0.9$, $\beta_2 = 0.999$, $\epsilon = 1\mathrm{e}{-7}$ \\

\textbf{Nadam} learning rate $\alpha = 0.001$, $\beta_1 = 0.9$, $\beta_2 = 0.999$, $\epsilon = 1\mathrm{e}{-7}$ \\

\textbf{RMSprop} learning rate $\alpha = 0.001$, $\rho = 0.9$, momentum$=0.0$, 
$\epsilon = 1\mathrm{e}{-7}$ \\

\textbf{Gravilon} none \\

\begin{table}[h]
\centering
\begin{tabular}{|l|c|c|}
\hline
 \text{Method} & \text{Accuracy} & \text{Best Accuracy}  \\ \hline
 Adagrad $\alpha=0.1$ & $0.976227$ & $0.9794$  \\ \hline
 Gravilon & $0.975487$ & $0.9785$ \\ \hline
 Nadam & $0.974547$ & $0.976$ \\ \hline
 Adam & $0.97398$ & $0.9753$ \\ \hline
 RMSprop & $0.969853$ & $0.9712$ \\ \hline 
 Adagrad $\alpha=0.01$ & $0.946487$ & $0.9477$  \\ \hline
 Adamax & $0.9454$ & $0.9472$ \\ \hline
 SGD & $0.927413$ & $0.9298$ \\ \hline
 Adagrad $\alpha=0.001$ & $0.75642$ & $0.7793$  \\ \hline
\end{tabular}
\caption{\label{tab:Trial1}The average classification accuracy of each gradient descent algorithm on the MNIST data set over $15$ trials with $2500$ steps in each trial.}
\end{table}

We find that Gravilon is a viable gradient descent algorithm for neural networks. In this example, we see that the Gravilon method provided the second highest average classification accuracy. Furthermore, only $0.074\%$ accuracy separated the first place Adagrad method, with learning rate $\alpha = 0.1$, and the Gravilon method. \\

In a subsequent experiment, for each method we trained the neural network until it achieved $95\%$ training accuracy. We utilized the same architecture for the neural network as in the previous experiment. For each method, we conducted $15$ trials, tracking the number of steps needed to achieve the target training accuracy and the resulting test accuracy. In Table \ref{tab:Trial2} we present the averaged results for each gradient descent algorithm. Additionally, we provide the best classification accuracy on the test data for each method over the $15$ trials. \\

\begin{table}[!h]
\centering
\begin{tabular}{|l|c|c|c|}
\hline
 \text{Method} & \text{Steps} & \text{Accuracy} & \text{Best Accuracy} \\ \hline
 Nadam & $110$ & $0.909713$ & $0.9216$ \\ \hline
 Adagrad $\alpha=0.1$ & $146.93$ & $0.906993$ & $0.919600$  \\ \hline
 Adam & $174.07$ & $0.92314$ & $0.930$ \\ \hline
 Gravilon & $325.87$ & $0.912953$ & $0.9306$ \\ \hline
 RMSprop & $376.07$ & $0.910907$ & $0.918600$\\ \hline 
 Adagrad $\alpha=0.01$ & $459.53$ & $0.907313$ & $0.915400$ \\ \hline
 Adamax & $864.4$ & $0.908113$ & $0.9149$ \\ \hline
 SGD & $1240$ & $0.89934$ & $0.91$ \\ \hline
 Adagrad $\alpha=0.001$ & $13492.8$ & $0.88713$ & $0.9003$  \\ \hline
\end{tabular}
\caption{\label{tab:Trial2}The average number of steps to achieve $95\%$ training accuracy over $15$ trials on the MNIST data set.}
\end{table}

In this second experiment, we find that Gravilon is more efficient than the SGD method. Additionally, the Gravilon method provides the second highest average classification accuracy on the test data. 

\subsection{CIFAR-10 Image Classification}
We implemented Gravilon in Python using the SGD optimizer in PyTorch. Our implementation follows the pseudocode shown in Algorithm \ref{alg:Gravilon}; we used the existing framework of the SGD optimizer and set the learning rate equal to $\beta$. In these experiments, we compared the abilities of three gradient descent algorithms (Adam, SGD, and Gravilon) in classifying the CIFAR-10 image dataset \cite{krizhevsky}. We relied on existing code to train our models. We only replaced the default optimizer with our optimizer and kept everything else the same. \\

In all of our experiments, we used a batch size of 128. Additionally, we made several adjustments to Gravilon to improve performance. We incorporated momentum, L2 regularization, and introduced a scaling parameter to increase the step size. For the settings, we used $0.9$ for momentum, $5\times 10^{-4}$ for weight decay, and rescaled the constant $\beta$ by $50$. For SGD, we used the default learning rate of $0.1$, momentum of $0.9$, and weight decay of $5\times 10^{-4}$. For Adam, we used the default hyperparameter values provided in the PyTorch documentation. We conducted experiments on three different neural network architectures: PreActResNet18 \cite{hePreAct}, ResNet18 \cite{heResNet}, and VGG19 \cite{simonyan}. 

\begin{figure}[!ht]
\includegraphics[scale=0.5]{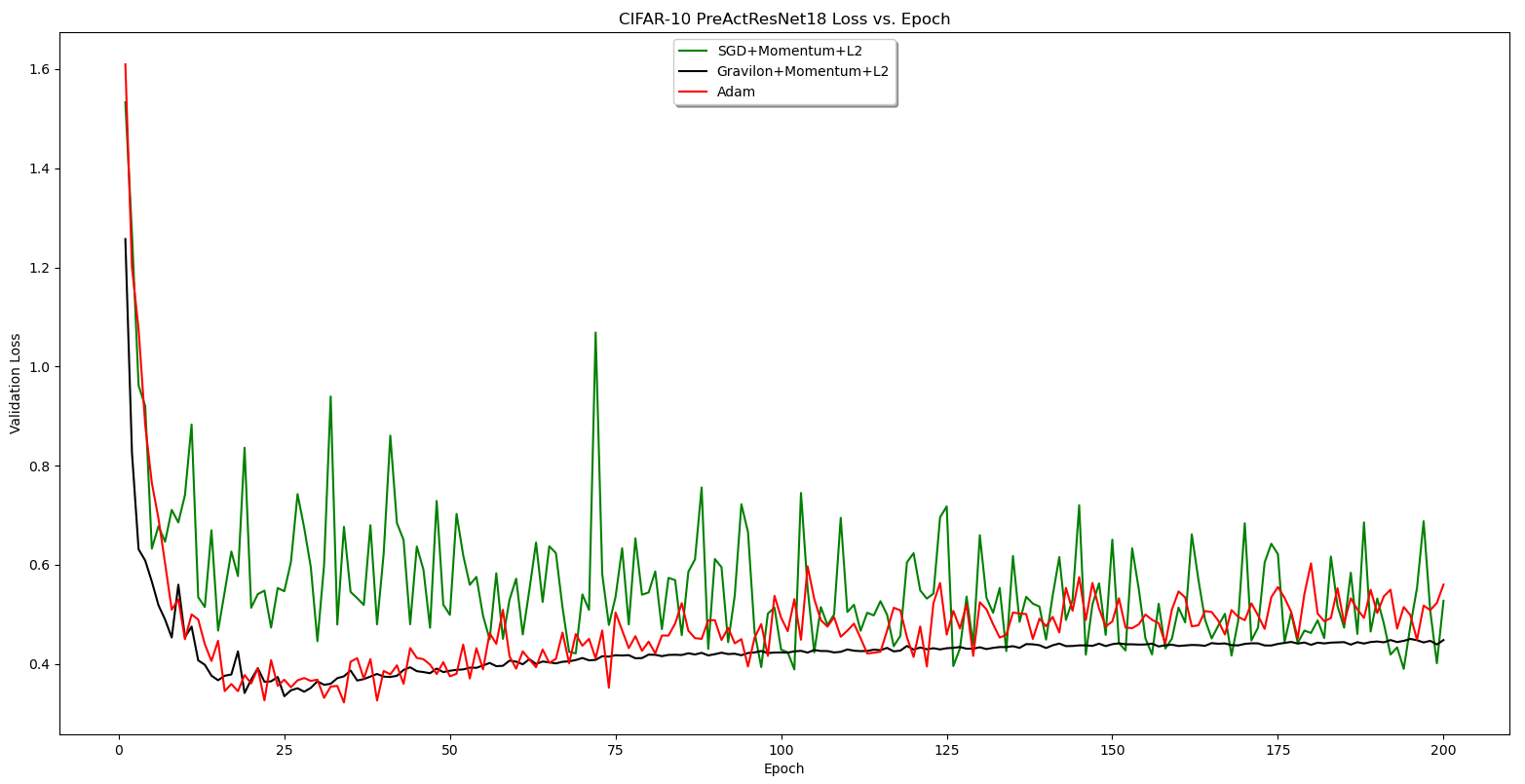}
\centering
\caption{Validation loss of PreActResNet18 network on CIFAR-10.}
\label{fig:PreAct}
\end{figure}

\newpage

\begin{figure}[!ht]
\includegraphics[scale=0.5]{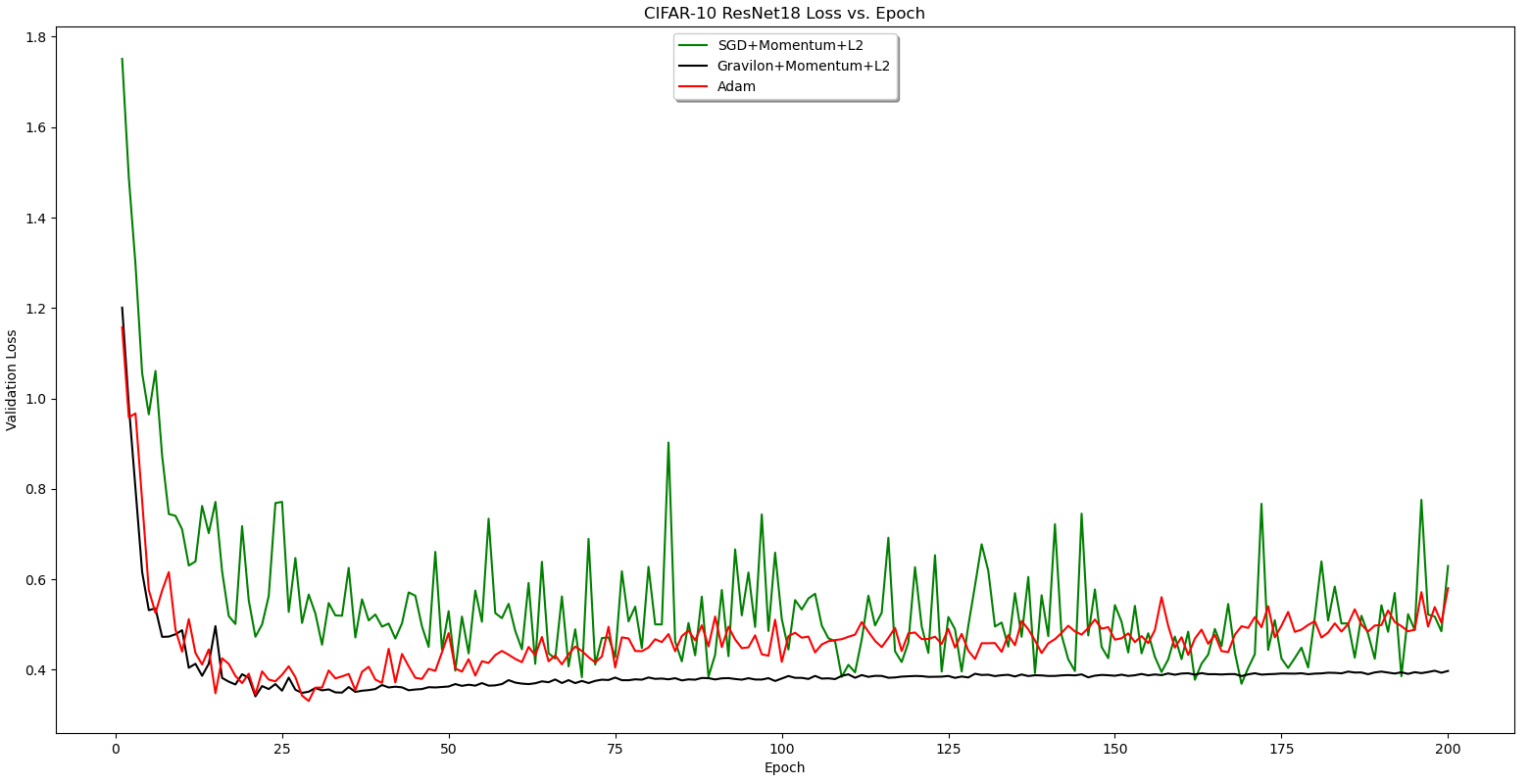}
\centering
\caption{Validation loss of ResNet18 network on CIFAR-10.}
\label{fig:ResNet}
\end{figure}

\begin{figure}[!ht]
\includegraphics[scale=0.5]{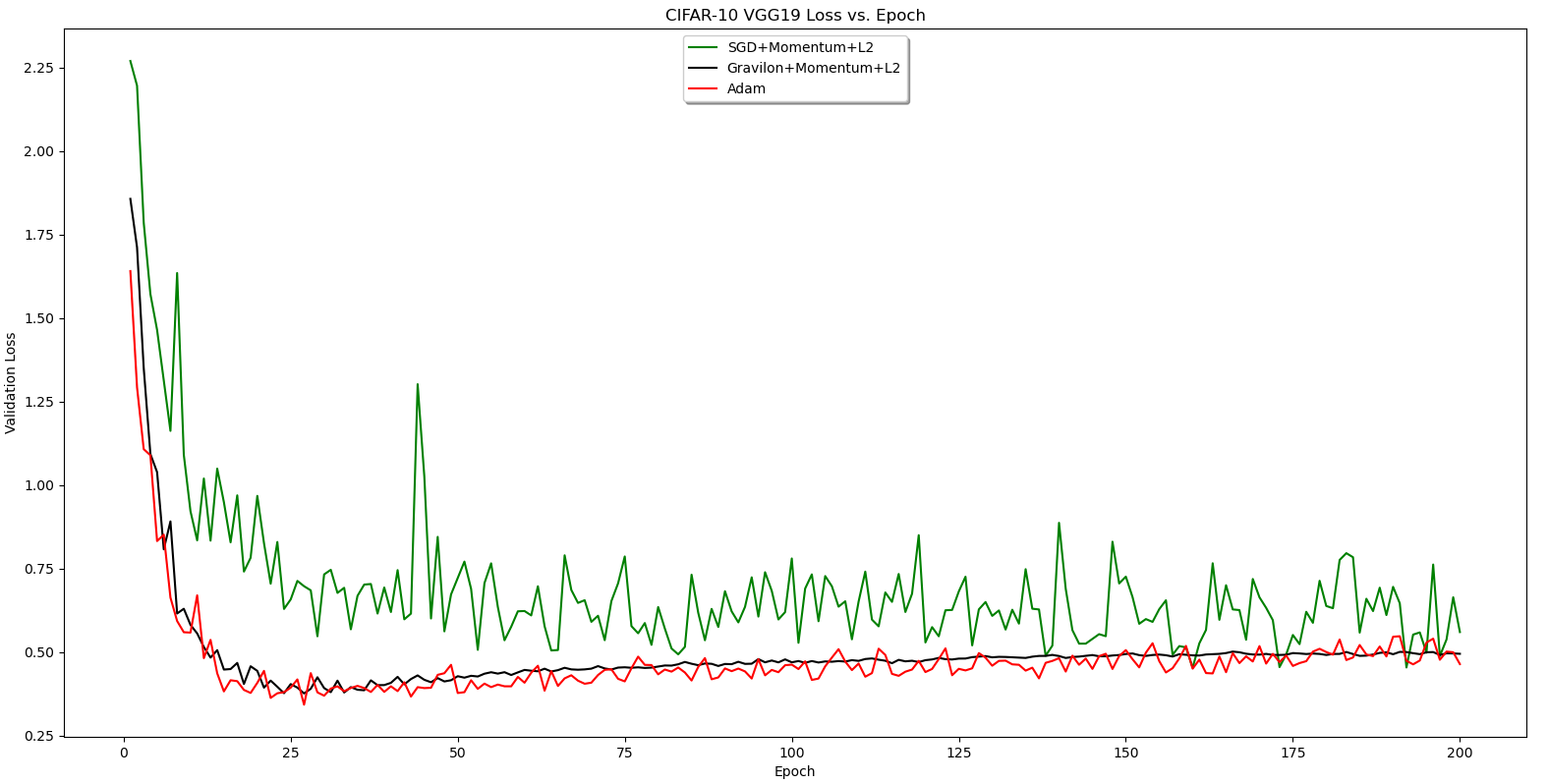}
\centering
\caption{Validation loss of VGG19 network on CIFAR-10.}
\label{fig:VGG}
\end{figure}

For the PreActResNet18 network, Figure \ref{fig:PreAct}, Gravilon achieved a minimum validation loss of $0.3351$ at epoch $25$, and Adam achieved a minimum loss of $0.3225$ at epoch $34$. For the ResNet18 network, Figure \ref{fig:ResNet}, Gravilon achieved a minimum validation loss of $0.341$ at epoch $21$, and Adam achieved a minimum loss of $0.3308$ at epoch $29$. For the VGG19 network, Figure \ref{fig:VGG}, Gravilon achieved a minimum validation loss of $0.3758$ at epoch $27$, and Adam achieved a minimum loss of $0.3421$ at epoch $27$.

\section{Conclusion}
Above, we presented a new gradient descent algorithm, Gravilon. Using the MNIST handwritten digit data set, we compared the resulting classification accuracy of the Gravilon method against the gradient descent algorithms SGD, Adagrad, Adam, Adamax, Nadam, and RMSprop. We showed experimentally that the Gravilon method provided the second best classification accuracy, second to only the Adagrad gradient descent algorithm. Additionally, we showed that the Gravilon method is more efficient than SGD in attaining $95\%$ training accuracy. \\

We believe that the Gravilon method, at the very least, should be considered as a replacement for the standard SGD algorithm. Experiments have consistently shown that Gravilon performs more efficiently and with a higher classification accuracy than standard SGD. Essentially, implementing the Gravilon method comes down to the addition of two lines of code. It is worth noting that the Gravilon method can be deployed concurrently with each of the gradient descent methods that we looked at in this experiment, which hints at the existence of a hybrid method further improving the efficiency and accuracy of the Gravilon method. \\

\section{Future Developments}
One of our next steps is to explore the effectiveness of the Gravilon method in deep learning applications such as in convolutional networks and recurrent networks. Additionally, as we work to develop a thorough theoretical understanding of the Gravilon method, we look to prove the convergence of the method in particular examples stemming from applications.

\end{document}